# Optimal Junction Trees

Finn V. Jensen    Frank Jensen
Department of Mathematics and Computer Science
Aalborg University
Fredrik Bajers Vej 7E, DK-9220 Aalborg Øst, Denmark
E-mail: fvj@iesd.auc.dk, fj@iesd.auc.dk

## Abstract

The paper deals with optimality issues in connection with updating beliefs in networks. We address two processes: triangulation and construction of junction trees. In the first part, we give a simple algorithm for constructing an optimal junction tree from a triangulated network. In the second part, we argue that any exact method based on local calculations must either be less efficient than the junction tree method, or it has an optimality problem equivalent to that of triangulation.

## 1 INTRODUCTION

The junction tree propagation method (Jensen et al., 1990; Lauritzen and Spiegelhalter, 1988) is designed for propagation in *Markov networks*:

- an undirected graph with discrete variables as nodes;

- for each clique U in the graph there is a potential $\phi_U$, which is a non-vanishing function from the set of configurations of U to the set of non-negative reals.

The *compilation* part of the method is to

- triangulate the graph (i.e., add extra links such that every cycle of length greater than three has a chord);

- form a potential $\phi_U$ for each clique U of the triangulated graph;

- construct a junction tree over the cliques.

A junction tree over the cliques is characterized by the so-called *junction tree property*: For each pair U, V of cliques with intersection S, all cliques on the path between U and V contain S.

The *propagation* part of the method consists of

- giving all links in the junction tree a *label* consisting of the intersection of the adjacent nodes; these labels are called *separators* (see Figure 1a);

- attaching a potential to all separators (initially the neutral potential consisting of ones);

- letting the nodes communicate via the separators: a message from U to V with separator S has the form that $\phi_U$ is marginalized down to S, resulting in $\phi'_S$; $\phi'_S$ is placed on the separator and $\phi'(S)/\phi(S)$ is multiplied on $\phi_V$ (see Figure 1b).

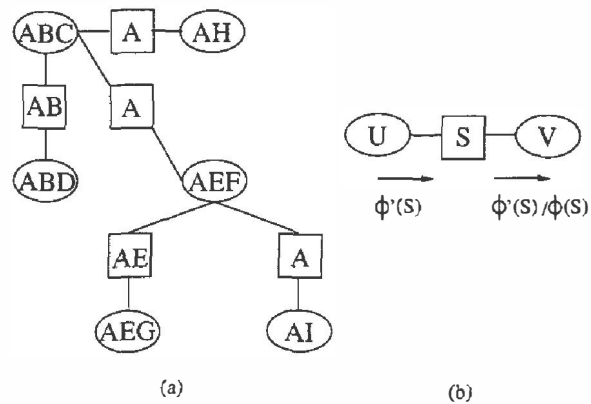

FIGURE 1.
(a) A junction tree. (b) Message passing in junction trees.

It is so, that after a finite number of message passes between neighbours in the junction tree, each potential in the junction tree holds the (possibly non-normalized) marginal of the joint probability distribution for the entire set of variables. In fact, the message passing can be organized so that it is sufficient with exactly one pass in each direction of the links in the junction tree. Therefore, in complexity considerations for propagation in junction trees, one can associate a local measure C(U, V) to links (U, V), where C(U, V) indicates time/space consumption for the two passes.

The compilation is not deterministic. Markov networks may have several different triangulations yielding different sets of cliques, and a triangulated network may have several different junction trees. We therefore would like to have algorithms yielding optimal triangulations and optimal junction trees with respect to complexity. However, the optimality problem for triangulations is $\mathcal{NP}$-complete (Arnborg et al., 1987).

In the first part of the paper, we address the optimality problem for junction trees given the triangulated graph, and we present a simple algorithm which is quadratic in the number of cliques.

In the last section, we address the triangulation process and ask the question whether it may be possible to come up with a propagation method which does not contain an $\mathcal{NP}$-hard optimality problem. The answer is discouraging. We show that any local calculation method must involve a hidden triangulation, and we use this to conclude that the method is either less efficient than the junction tree method, or it has an $\mathcal{NP}$-hard optimality problem.

## 2 JUNCTION TREES AND MAXIMAL SPANNING TREES

Throughout the remainder of the paper, we consider a triangulated connected graph G with clique set $\mathcal{C}$. The cliques of G are denoted by the letters U, V, W, U', etc. We shall not distinguish between a clique and its set of variables. So we talk of the intersection of cliques meaning the set of variables common to the cliques. Intersections are denoted by letters R, S, R', etc.

**Definition 1** The *junction graph* for G has $\mathcal{C}$ as nodes, and for each pair U, V of cliques with nonempty intersection R there is a link with label R. Each link has a *weight* which is the number of variables in the label.

**Theorem 1** *A spanning tree for the junction graph of G is a junction tree if and only if it is a spanning tree of maximal weight.*

Theorem 1 has been proved independently by Shibata (1988) and Jensen (1988). Here we will give a proof much simpler than the original ones. Before giving the proof, we shall recall two algorithms for the construction of maximal spanning trees.

**Algorithm 1 (Prim)**
(1) Put $\mathcal{N} = \{U\}$, where U is an arbitrary node.
(2) Choose successively a link $\{W, V\}$ of maximal weight such that $W \in \mathcal{N}$ and $V \notin \mathcal{N}$, and add V to $\mathcal{N}$.

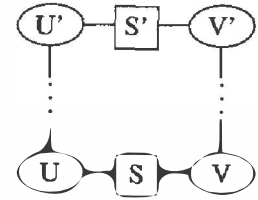

**FIGURE 2.** Paths in T and T'.

Prim's algorithm constructs a sequence $T_0 \subseteq \cdots \subseteq T_n$ of maximal spanning trees for the subgraph determined by $\mathcal{N}$.

**Algorithm 2 (Kruskal)**
Choose successively a link of maximal weight not producing a cycle.

Kruskal's algorithm works with a forest of partial maximal weight spanning trees. Whenever a link is chosen, two partial trees are connected into a new partial spanning tree of maximal weight.

Both algorithms result in maximal weight spanning trees, and each maximal weight spanning tree can be constructed through any of the two algorithms. [Proofs can be found in many textbooks on graph algorithms, e.g., (Goudran and Minoux, 1984) and (McHugh, 1990)].

*Proof of Theorem 1:* Let T be a spanning tree of maximal weight. Let it be constructed by Prim's algorithm such that $T_1 \subseteq \cdots \subseteq T_n = T$ is a sequence of partial maximal weight spanning trees.

Assume that T is not a junction tree. Then, at some stage m, we have that $T_m$ can be extended to a junction tree T' while $T_{m+1}$ cannot. Let (U, V) with label S be the link chosen at this stage; $V \in T_{m+1}$ (see Figure 2).

Since $T_{m+1}$ cannot be extended to a junction tree, the link (U, V) is not a link in T'. So, there is a path in T' between U and V not containing (U, V). This path must contain a link (U', V') with label S' such that $U' \in T_m$ and $V' \notin T_m$ (see Figure 2).

Since T' is a junction tree, we must have $S \subseteq S'$, and since S was chosen through Prim's algorithm at this stage, we also have $|S| \geq |S'|$. Hence, $S = S'$.

Now, remove the link (U', V') from T' and add the link (U, V). The result is a junction tree extending $T_{m+1}$, contradicting the assumption that it cannot be extended to a junction tree.

Next, let T be any non-maximal spanning tree. We shall prove that T is not a junction tree. Again, let $T_1 \subseteq \cdots \subseteq T'$ be a sequence of maximal trees con-structed through Prim's algorithm. Let the construction be so that a link from T is chosen whenever possible. Let $m$ be the first stage where this is not possible, and let $(U, V)$ with separator S be the link actually chosen $(U \in T_m, V \notin T_m)$. In T there is a path between U and V. As in the first part of the proof, we have that this path contains a link $(U', V')$ with label $S'$ such that $U' \in T_m$ and $V' \notin T_m$ (see Figure 2). Since $(U', V')$ could not be chosen, we have $|S'| < |S|$, and therefore S contains variables not in $S'$. Hence, T does not satisfy the junction tree condition. ∎

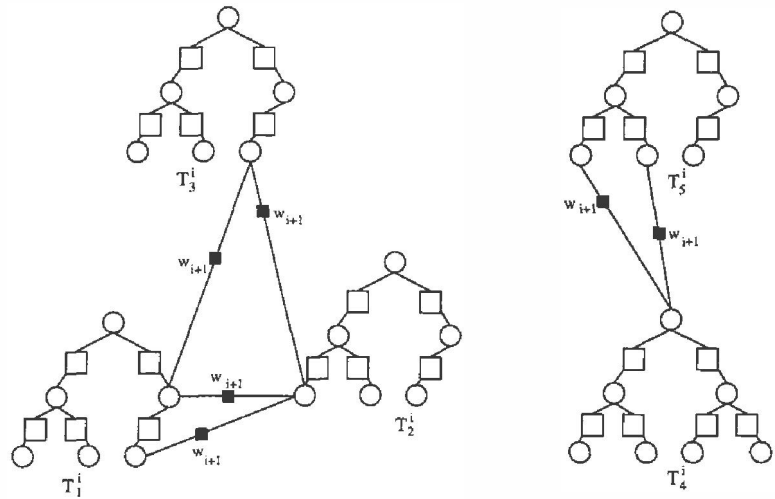

FIGURE 3. The thinning task at stage $i+1$ in Kruskal's algorithm.

## 3 OPTIMAL JUNCTION TREES

Whenever the junction graph has several spanning trees of maximal weight, there are accordingly several junction trees. Assume that there is a real-valued measure on junction trees yielding a priority among them, and assume that this measure can be decomposed to a local measure $C(U, V)$ attached to the links. We call the measure a *cost*. We may also assume that the entire measure is strictly increasing in the local measures, and that an optimal junction tree is one of minimal cost.

Let us take a closer look at the construction of junction trees through Kruskal's algorithm. Let $w_1, \ldots, w_n$ be the different weights of G in decreasing order. The algorithm can be considered as running through $n$ stages characterized by the weight of the links chosen. At the end of stage i, all links possible of weight $w_1, \ldots, w_i$ have been chosen, and a forest $T_1^i, \ldots, T_{k_i}^i$ of partial maximal weight spanning trees has been constructed.

Now, the task at stage $i+1$ can be considered in the following way. Add all links of weight $w_{i+1}$ to the forest, and break the cycles by removing links of weight $w_{i+1}$. Any thinning will result in a forest of partial spanning trees of maximal weight. Note that any thinning at a given stage will result in the same connected components, and therefore the thinning chosen has no impact on the next stage. Hence, if we in the construction have a secondary priority (cost, say), we can perform the thinning by using Kruskal's algorithm according to cost. In this way we will end up with a maximal weight spanning tree of minimal cost (see Figure 3).

We conclude these considerations with

**Theorem 2** *Any minimal cost junction tree can be constructed by successively choosing a link of maximal weight not introducing cycles, and if several links may be chosen then a link of minimal cost is selected.*

A proof of Theorem 2 is an induction proof over the stages. The induction hypothesis is that at the end of each stage, the forest consists of partial maximal distance junction trees.

**Remark 1** *An analoguous algorithm based on Prim's algorithm will also construct minimal cost junction trees.*

**Corollary 1** *All junction trees over the same triangulated graph have the same separators (also counting multiplicity).*

*Proof:* Consider stage $i+1$ (Figure 3). A cycle can be broken by removing any link of weight $w_{i+1}$. If $(U, V)$ with separator S is removed, then all separators in the remaining paths between U and V must contain S. This means that any separator of weight $w_{i+1}$ on these paths must equal S. By thinning we therefore remove the same separators. ∎

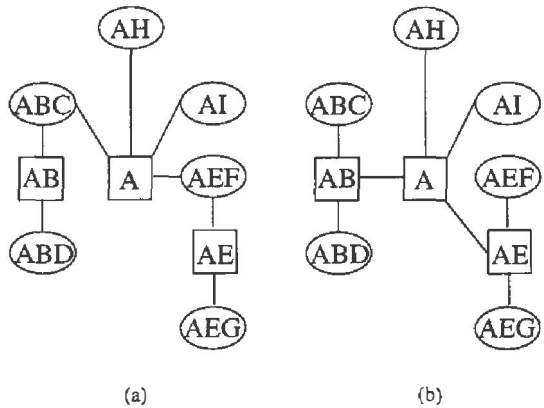

**FIGURE 4.**
(a) Contraction of the junction tree from Figure 1.
(b) An Almond tree.

## 4 ALMOND TREES

Almond and Kong (1993) suggest another type of junction tree. Compared to the junction trees in (Jensen et al., 1990), they give some reduction in computational complexity.

**Observation 1** If $n$ links have the same separator, the communication scheme can be contracted (Figure 4a).

In junction trees, each separator holds exactly one potential table where the marginal last communicated is stored. In contracted junction trees, a separator with $n$ neighbours must hold at least $n - 1$ potential tables to store marginals communicated from neighbours. This means that there is no saving in space. There is, however, a saving in time, since a number of marginalizations are avoided.

**Observation 2** If a separator is a subset of another separator, they can be linked (Figure 4b).

The type of calculations are the same for links between separators as for links between separators and cliques. Due to the corollary, we know for each separator $S$, the number of supersets to which it shall be linked, and for each link $(S, S')$, we can associate a local cost $C(S, S')$.

Junction trees simplified through these two observations we call *Almond trees*. The construction of an Almond tree may go as follows:

- From the triangulated graph, the set of cliques and the set of separators (including multiplicity) is established. This can be done through elimination in the triangulated graph, but it is not important for our considerations.

- For each separator, establish links to all cliques and separators containing it.

- For each separator (with multiplicity $n$), choose $n+1$ links to supersets without introducing cycles.

**Theorem 3** *Any minimal cost Almond tree can be constructed by successively choosing links for separators of maximal weight, and if several links may be chosen, take one of minimal complexity.*

A proof of Theorem 3 is an induction proof along the same line as a proof of Theorem 2.

## 5 THE NECESSITY OF TRIANGULATION

In the former sections we gave an efficient algorithm for constructing optimal junction trees given the triangulated graph. Thereby all steps from DAG to junction tree is covered by efficient algorithms yielding an optimal output — except for the triangulation. Since this problem is $\mathcal{NP}$-complete, we cannot hope for an efficient algorithm yielding an optimal triangulation. It appears that a one-step look-ahead heuristic provides the best triangulations. An alternative propagation scheme is *conditioning* (Pearl, 1988). The $\mathcal{NP}$-complete part of conditioning is the determination of a cut set for the DAG, and Becker and Geiger (1994) have given an algorithm which guarantees a cut set space no larger than the square of the space for an optimal cut set. Other schemes exist, like, e.g., arc-reversal (Shachter, 1990); however, as has been shown by Shachter et al. (1991), all known methods do in fact contain a hidden triangulation.

Since belief updating in Bayesian networks is $\mathcal{NP}$-hard (Cooper, 1990), there is not much hope of finding a scheme avoiding an $\mathcal{NP}$-hard step. However, Cooper's result does not yield that any scheme will contain such a step. Cooper showed that through belief updating, the satisfiability problem for propositional calculus can be solved, but it may still be so that a search for an optimal structure for belief updating is polynomially solvable. Note namely that the space of the cliques are exponential in their presentation.

Also, new schemes are proposed (Zhang and Poole, 1992) which may seem as if they avoid the triangulation problem. We will in this section argue that *any* scheme for belief updating — meeting certain requirements — will contain a hidden triangulation. Then, if the complexity ordering of the hidden triangulations follows the ordering in the original scheme, we can conclude that if the scheme has a polynomially solvable optimality problem, then the junction tree method either provides more efficient solutions or $\mathcal{P} = \mathcal{NP}$.

The considerations to come are somewhat speculative and at places they need further precision. Hence, we call the results 'statements' rather than theorems. However, a reader looking for alternative propagation methods can use them as guidelines preventing investigations of several alternatives.

### Specifications

$U = \{A, \ldots, B\}$ is a *universe* consisting of a finite set of discrete variables. The joint probability $P(U)$ is a distribution over the configurations $\mathcal{X}_U = A \times \cdots \times B$.

A *local representation* of $P(U)$ consists of a set $\{P(U_1), \ldots, P(U_n)\}$, where $U_1, \ldots, U_n$ is a covering of $U$, and $P(U_i)$ is the marginal distribution of $U_i$.

A local representation can be visualized by a graph $G$ with the variables as nodes and with a link between two variables if there is a $U_i$ containing both; $G$ is called the *representing graph*.

The propagation task can be formulated as follows. Let $P'(U_i)$ be substituted for $P(U_i)$; if $P'(U) = P(U) \times P'(U_i)/P(U_i)$ is well-defined, then calculate the new marginals $P'(U_1), \ldots, P'(U_n)$.

By a *scene* for a propagation task, we understand a universe $U$ together with a covering $U_1, \ldots, U_n$ such that the covering equals the cliques in the representing graphs. An *instance* of a propagation task is a pair $(G, \mathcal{P})$, where $G$ is an undirected graph, and $\mathcal{P}$ is a set of marginals of a joint distribution $P(U)$ to the cliques of $G$.

Let $U$ be a universe. By a *local method* on $U$, we understand an algorithm working only on subsets of $U$. More precisely: The algorithm consists of a control structure and a fixed set $Pr_1, \ldots, Pr_m$ of procedures such that each $Pr_i$ only processes information on $V_i \subseteq U$. We call $V_i$ the *scope* of $Pr_i$. The *representing graph* $G'$ for a local method is defined as the graph with $U$ as nodes, and with links between variables if there is a scope containing them. Notice that the cliques of $G'$ need not be scopes.

We have defined a local method such that the control structure mainly consists of controlling message passing between procedures. Note that between $Pr_i$ and $Pr_j$ only information on $V_i \cap V_j$ is worth passing.

A *general local belief updating method* for a scene represented by $G$ is a local method solving the propagation task for each instance $(G, \mathcal{P})$.

We aim at the following:

**Statement 1** *Let $G$ represent a scene, and let a general local belief updating method be represented by the graph $G'$. Then $G'$ contains a triangulation of $G$.*

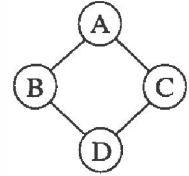

**FIGURE 5.**
A graph representing a general propagation task.

First, we shall transform the problem to propositional calculus.

**Lemma 1** *Let $P(U_1), \ldots, P(U_m)$ be projections of the joint probability table $P(U)$. Let $Pos(U)$ be the table of possible configurations of $U$:*

$$Pos(u) = \begin{cases} 1 & \text{if } P(u) > 0 \\ 0 & \text{otherwise} \end{cases}$$

*Define $Pos(U_i)$ as:*

$$Pos(u_i) = \begin{cases} 1 & \text{if } P(u_i) > 0 \\ 0 & \text{otherwise} \end{cases}$$

*Then $Pos(u_i) = 1$ if and only if $u_i$ is a projection of a possible configuration.*

*Proof:* Since $P(U_i)$ is the marginal of $P(U)$, we have that $P(u_i) > 0$ if and only if $u_i$ is the projection of at least one configuration with positive probability. ∎

The lemma shows that any scheme for belief updating has the calculus of possible configurations in propositional calculus as a special case. So, if we can prove Statement 1 for this calculus, we are done.

We shall start with an example which is the cornerstone of the proof.

**Example 1** Let the graph in Figure 5 represent a general propagation task over the propositional calculus, and let $Pos$ be the potential giving 1 for possible configurations and 0 for impossible ones.

Let $Pr_{AB}, Pr_{AC}, Pr_{BD}, Pr_{DC}$ be procedures for solving the task (the index indicates the scope, see Figure 6).

We shall construct an instance which cannot be solved by the procedures. For each variable we only use the first two states. This means that all other states are impossible.

Initially, we have for $i, j \leq 2$

$$\begin{array}{ll} Pos(a_i, b_j) = 1 & \text{for all } i, j \\ Pos(a_i, c_j) = 1 & \text{if and only if } i = j \\ Pos(b_i, d_j) = 1 & \text{if and only if } i = j \\ Pos(c_i, d_j) = 1 & \text{for all } i, j \end{array}$$

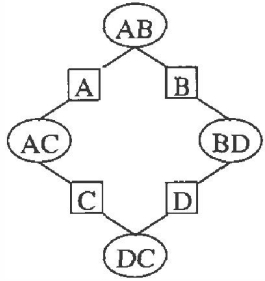

**FIGURE 6.**
The scopes for the procedures and the communication channels.

That is, A and C as well as B and D are forced into the same state, and everything else is possible. Note that the Pos-relations above are projections of the Pos-relation over the universe:

$$Pos(a_i, b_j, c_k, d_\ell) = 1$$

if and only if

$$Pos(a_i, b_j) = Pos(a_i, c_k)$$
$$= Pos(b_j, d_\ell) = Pos(c_k, d_\ell) = 1$$

Now, assume we get the information that the configurations $(a_1, b_2)$ and $(a_2, b_1)$ are impossible. This is equivalent to replacing the relation $Pos(a_i, b_j)$ by

$$Pos'(a_i, b_j) = 1 \quad \text{if and only if} \quad i = j \quad (\text{and } i, j \leq 2).$$

Now, the propagation task is to determine $Pos'(A, C)$, $Pos'(B, D)$, and $Pos'(C, D)$ such that these local relations are projections of the unique universal relation $Pos'(A, B, C, D)$, satisfying the relations $Pos'(A, B)$, $Pos(A, C)$, $Pos(B, D)$, and $Pos(C, D)$.

Clearly, $Pos'(a_i, b_j, c_k, d_\ell) = 1$ if and only if $i = j = k = \ell$, and therefore $Pos'(c_k, d_\ell) = 1$ if and only if $k = \ell$.

The tool for achieving this result is the set $Pr_{AB}$, $Pr_{AC}$, $Pr_{BD}$, and $Pr_{CD}$ of procedures. Since $Pr_{AB}$ can only process information on the variables A and B, and $Pr_{AC}$ can only process information on A and C, then the only valuable information to communicate between the two procedures is information on A (see Figure 6). That is, between $Pr_1$ and $Pr_2$ with scopes $V_1$ and $V_2$, respectively, only information on $V_1 \cap V_2$ need to be communicated. The new relation $Pos'(A, B)$ introduces a constraint between the state of A and the state of B, but since only information on A alone and B alone can be communicated, the constraint cannot be communicated to $Pr_{CD}$.

Note that if a cycle contains more than 4 variables, the construction can be extended by clamping the states of further intermediate variables.

*Proof of Statement 1:* Assume that $G'$ does not contain a triangulation of G. Then there is a cycle C in G such that the subgraph of $G'$ consisting of the nodes in C is not triangulated. Let $C'$ be a chordless cycle of length greater than three in that subgraph. Let $A_1, \ldots, A_n$ be the nodes of $C'$.

We now can construct an instantiation, which cannot be propagated correctly: (1) Let a configuration be possible if and only if its projection to $A_1 \times \cdots \times A_n$ is possible. (2) Perform the construction as shown in the example. ∎

By the proof of Statement 1, we see that it can be generalized to systems with other uncertainty calculi like, e.g., Dempster-Shafer belief functions or fuzzy systems. In fact, the reasoning can be applied to any calculus having propositional calculus as a special case. An axiomatization of these possible calculi is outside the scope of this paper, but the axioms in (Shenoy and Shafer, 1990) form a good starting point.

Concerning complexity we still have a couple of loose ends. Although a general scheme involves a hidden triangulation, the computational complexity needs not be of the same kind as for the junction tree scheme. In the junction tree scheme the complexity is proportional to the number of configurations in the cliques. Therefore a general local scheme has an equivalent computational complexity if it is proportional to the number of configurations in the scopes. This is the case if each configuration has an impact on the messages sent in the algorithm. In this paper we shall not give sufficient conditions for this to hold.

The second loose end has to do with optimality. A general scheme is, e.g., to work with $P(U)$ only. This corresponds to working with the complete graph over $U$. This scheme has a trivial optimality problem, but the junction tree method can do much better even for suboptimal triangulations. Therefore we conclude:

**Statement 2** *If a general local propagation scheme has a complexity at least proportional to the number of configurations in the scopes, and its optimality problem can be solved in polynomial time, then either the junction tree scheme can do better or $\mathcal{P} = \mathcal{NP}$.*

### Acknowledgements

The work is part of the ODIN-project at Aalborg University, and we thank our colleagues in the group for inspiring discussions.

The work is partially funded by the Danish Research Councils through the PIFT-programme.